\documentclass[12pt]{article}

\usepackage{amsmath,amssymb,amsthm,mathtools,dsfont}
\usepackage{url, graphicx, enumitem, setspace, subcaption, multirow}
\usepackage{bm}
\usepackage{soul}
\usepackage{authblk}
\usepackage[margin=1in]{geometry}
\usepackage{listings}
\usepackage{xcolor, booktabs, array}
\usepackage{caption, float}

\begin{document}
\singlespacing
\title{Fire-EnSF:  Wildfire Spread Data Assimilation using \\Ensemble Score Filter}
\author[1]{Hong Zheng Shi}
\author[2]{Yuhang Wang}
\author[1]{Xiao Liu}
\affil[1]{H. Milton Stewart School of Industrial and Systems Engineering}
\affil[2]{School of Earth and Atmospheric Sciences

Georgia Institute of Technology}
\date{}
\maketitle

\begin{abstract}
As wildfires become increasingly destructive and expensive to control, effective management of active wildfires requires accurate, real-time fire spread predictions. To enhance the forecasting accuracy of active fires, data assimilation plays a vital role by integrating observations (such as remote-sensing data) and fire predictions generated from numerical models. 
This paper provides a comprehensive investigation on the application of a recently proposed diffusion-model-based filtering algorithm---the Ensemble Score Filter (EnSF)---to the data assimilation problem for real-time active wildfire spread predictions. Leveraging a score-based generative diffusion model, EnSF has been shown to have superior accuracy for high-dimensional nonlinear filtering problems, making it an ideal candidate for the filtering problems of wildfire spread models. Technical details are provided, and our numerical investigations demonstrate that EnSF provides superior accuracy, stability, and computational efficiency, establishing it as a robust and practical method for wildfire data assimilation. Our code has been made publicly available. 


\end{abstract}

\section{Introduction}

Over the past decades, extended droughts, heat waves, and increases in surface and air temperatures have created more favorable conditions for ignition and fire spread in the United States, even in areas that would not normally burn. Since 2010, more than 93,000 structures have been destroyed by wildfires in the U.S. \cite{nicc}, and eighteen of the twenty most destructive fires in California occurred during this time period \cite{CALFIREdata}.  The recent Palisades and Eaton fires in Los Angeles County are two of California's most destructive wildfires, which lasted over 24 days from Jan 7 to 31, 2025, destroyed over 15,000 structures, caused approximately 30 fatalities, and 10 firefighter injuries. The total containment cost exceeded $\$250$ million \cite{calfire_eaton_2025,calfire_palisades_2025}.

With wildfires becoming more destructive and expensive to control, effective management of active wildfires requires accurate real-time fire spread predictions. This critical capability assists fire agencies in determining how resources should be allocated and which areas should be evacuated. It is noted that the natural fire regime across the U.S.ecosystems has been disrupted by many factors, including fire exclusion, land management, and socio-economic issues. 
Such disturbances in the fire regime have contributed to the change in fuel type and distribution across the landscape, negatively affecting the accuracy of operational models used by fire agencies to manage resources and suppress wildfires \cite{North2015, Stephens2022}. Hence, data assimilation becomes a critical solution for accurate real-time fire predictions by fusing the output generated by numerical fire spread models and observation data streams. 

This paper investigates the application of a recently proposed diffusion-model-based filtering algorithm---the Ensemble Score Filter (EnSF) \cite{BAO2024, si2024latent}---to the data assimilation problem for numerical wildfire spread models. Leveraging a score-based diffusion model, EnSF has been shown to have superior accuracy for high-dimensional nonlinear filtering problems, making it an ideal candidate for the filtering problems of wildfire spread models. 
Throughout the paper, we focus on the FARSITE fire modeling and simulation model when investigating the application of EnSF for such problems (although the same approach can be applied to other fire spread models by replacing the FARSITE predictions with others in the framework described in this paper). Our code has been made available for readers. 

\subsection{An Overview of Wildfire Spread Models}
Wildfires are complex physical processes governed by the interaction of environmental processes across multiple scales. 
A physics-based numerical wildfire simulation model (e.g., WRF-SFIRE) couples a mesoscale weather model (e.g., Weather Research and Forecasting (WRF)) and a 2D fire spread model (e.g., SFIRE) \cite{Mandel2011,Clark2016,Coen2013,kochanski2013evaluation}. 
The coupling captures the interactions between weather and fire behaviors; e.g., wind affects fire spread, while fire affects atmospheric variables through the heat/vapor fluxes. 
Many of the fire spread models are 2D semi-empirical (e.g., the classical Rothermel model \cite{Rothermel1972, rothermel1983predict}) that consist of numerical algorithms for fireline propagation, fuel computation, and related functions. For example, the SFIRE model is based on the level set method and produces vector fire polygons \cite{Osher2003}. FARSITE, another widely used model by the U.S. Forest Service, is largely based on Rothermel's model and produces vector fire perimeter polygons where the vertices of these polygons contain information about the fire's spread rate and intensity and are propagated \cite{Finney2004}. In Section \ref{sec:FARSITE}, a more detailed review of FARSITE is provided. 

Numerical wildfire simulation models are highly interpretable and provide useful insights on wildfire spread under various atmospheric conditions, long-term forecasting and planning, ``what-if'' analysis, etc. On the other hand, such models are faced with some major challenges arising from real-time active fire prediction and management: 

(\textbf{\textit{i}}) For real-time operations and decisions, computational time becomes a major obstacle that prevents us from using numerical simulation models, which may take several hours for a single run \cite{Bottero2020}. 
(\textbf{\textit{ii}})
In addition, physics-based or semi-empirical wildfire simulation models require accurate inputs for a large number of parameters (e.g., fuel wind height, fuel roughness height, fuel depth, total fuel load, etc.) that are often never precisely known for real-time operations. 
(\textbf{\textit{iii}}) Due to the challenge above, data assimilation is often a critical step in improving the accuracy of active wildfire predictions by integrating simulation output and noisy observational data. However, because of the high computational cost, assimilating multi-source data streams that contain valuable real-time information on wildfire dynamics (e.g., geostationary remote-sensing data available, aerial observations from suborbital platforms, etc.) remains a significant challenge \cite{wei2024statistical}. 
(\textbf{\textit{iv}}) Most of these numerical simulators are deterministic, and it is a challenging task to perform uncertainty quantification that requires repeated simulation runs. As it is already noted in \cite{SRIVAS2016897}, models such as FARSITE are essentially deterministic and do not incorporate any stochastic components (even though the wildfire spread is a complicated natural process often subject to a high level of uncertainty). 

\vspace{6pt}
Over the past decade, we would also like to point out that Machine Learning (ML) approaches, such as Deep Learning, provide complementary strengths for active wildfire spread prediction \cite{Hodges2019, ijcai2019p636, Crowley2019, Huot2020}. 
The key idea is to predict the probability that the burn map reaches a spatial grid (image pixel) within a time window of interest, given environmental variables such as topography, wind, vegetation, drought index, and population density (i.e., a supervised binary classification problem). For example, \cite{Hodges2019} and \cite{ijcai2019p636} employed the Convolutional Neural Networks (CNN) to generate the burn map and the latter applied the approach to 2016 Beaver Creek fire in Colorado, \cite{Crowley2019} modeled the wildfire spread process as a Markov Decision Process based on deep reinforcement learning, and a team of researchers from Google recently used a Convolutional Autoencoder---a type of neural network for precise image segmentation--to predict if fire will reach a given image pixel on the next day \cite{Huot2020}. Although the off-line training of these ML models can still be computationally expensive, the on-line prediction is usually fast enough for real-time predictions. 
However, one limitation, which has received much attention in recent years, is the lack of explainability and connection to domain knowledge for many black-box ML approaches. It is important to note that fire management involves high-stakes decisions for which domain knowledge imposes critical constraints on how data should be modeled and how models can be interpreted. Stakeholders who make decisions and fight against wildfires on the front line may not be data scientists by training. As a result, the lack of explainable models and actionable insights has become the main barrier that significantly impedes the implementation of the latest advances in ML into wildfire science and management---an increasingly data-rich and domain-knowledge-intensive application domain. 

\subsection{Data Assimilation using EnSF}

Following the discussions above, this paper focuses on data assimilation for wildfire spread models by integrating simulation output and near real-time observations. In the literature, the Ensemble Kalman Filter (EnKF) and sequential Monte-Carlo approach (particle filters) have so far been the most widely adopted approaches. The use of EnKF not only enables the estimation of the latent system states (i.e., the coordinates of the fire perimeter) but also allows for systematic uncertainty quantification for the estimated coordinates of the fire perimeter. For example, \cite{SRIVAS2016897} provided a classical example of using EnKF for assimilating the FARSITE output and noisy finite spatial resolution measurements.  
To expedite data assimilation, \cite{yoo2023rapid} also adopted the EnKF and represented the wildfire perimeter by a two-dimensional polyline simplification algorithm. In addition, \cite{Ge2024} investigated the assimilation of UAV-based observation data for a discrete event wildfire spread simulation model using a particle filter-based data assimilation algorithm; also see \cite{xue2012data} for the use of sequential Monte Carlo methods in wildfire spread simulation.  Other approaches, such as the conditional Wasserstein generative adversarial networks trained with WRF–SFIRE simulations, have also been used to infer the fire arrival time from satellite active fire data \cite{shaddy2024generative}. 

\vspace{6pt}
A potential drawback of EnKF and particle filters is the low accuracy for high-dimensional nonlinear problems. 
In recent years, diffusion models have emerged as a powerful generative model with record-breaking performance in many applications, and an Ensemble Score Filter (EnSF) has been proposed for solving high-dimensional nonlinear filtering problems \cite{BAO2024, si2024latent}. The EnSF utilized a score-based diffusion model to effectively generate samples from the posterior distribution of the system state. 

In particular, three salient advantages of EnSF make it potentially useful and worth investigating for wildfire spread data assimilation:

(i) Unlike EnKF or particle filters, which require generating a large set of finite Monte Carlo samples, EnSF stores the information of the recursively updated filtering density in the score function, making it effective in handling nonlinear high-dimensional problems. 

(ii) Unlike existing diffusion models that require the training/tuning of neural networks, \cite{BAO2024} developed a training-free score estimation that directly approximates the score function, making EnSF computationally efficient for real-time wildfire spread data assimilation. 

(iii) As shown in \cite{si2024latent}, the EnSF are suitable for data assimilation with both dense and sparse observations. Both scenarios are relevant for wildfire spread problems. For example, remote-sensing data could be dense for a certain area, while UAV monitoring data are often sparse considering the spatial coverage of individual UAVs. 

\begin{figure}[h!]
    \centering
    \includegraphics[width=0.6\linewidth]{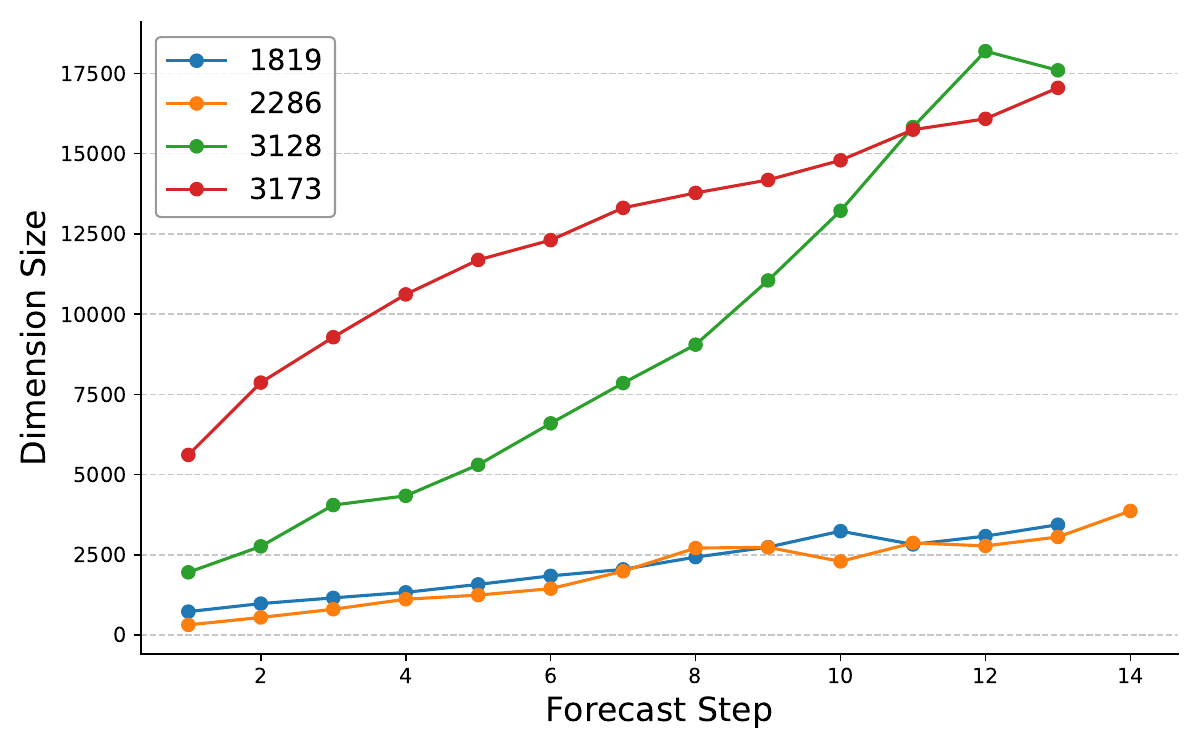}
    \caption{The growth of the number of vertices (of the fire boundary polygons) in FARSITE wildfire simulations (the legend on the top right shows the fire ID for four testing fire use cases)}
    \label{fig:dim}
\end{figure}
Finally, it is worth mentioning another special challenge associated with the data assimilation for wildfire spread models. Recall that many fire spread models, such as FARSITE, produce vector fire perimeter polygons
where the vertices of these polygons are propagated. As a result, the number of vertices in the predicted fire boundary polygon increases with simulation steps (as the burning area expands). In other words, the dimension of the system state vectors grows in time. 
As an illustration, Figure \ref{fig:dim} provides an example of how the number of vertices (of the fire boundary polygons) grows during the FARSITE wildfire spread simulation. In all four testing fire use cases, we could clearly see the growth of the number of vertices, indicating that the underlying dynamical model has a non-constant (i.e., increasing) dimension. This continual growth in the state dimension is known to significantly challenge the EnKF or particle filters to track the true fire perimeter accurately.


\vspace{6pt}
Hence, this paper presents a timely investigation (the first in the literature to our best knowledge) of leveraging the new EnSF for addressing the data assimilation problems for wildfire spread models. In particular, we focus on the FARSITE fire spread model (although the framework applies to many other simulation models). FARSITE is an operational fire growth simulation model that computes wildfire growth under heterogeneous conditions of terrain, fuels, and weather \cite{mbedwardFARSITE}. Sits in the core of FARSITE is the Rothermel spread model \cite{scott2005}, which uses a set of empirical equations derived from laboratory experiments to calculate fire spread rate based on complex spatial data on topography, fuel characteristics, and weather conditions, relying on standardized fuel models to characterize the vegetation that carries the fire \cite{scott2005}. Currently, the FARSITE module has been integrated into a comprehensive fire modeling system called FlamMap6 \cite{FlamMap}. 

\vspace{6pt}
The paper is organized as follows. Section \ref{sec:FARSITE} first provides some essential preliminaries on FARSITE, detailing its mathematical foundation, core algorithms, and operational characteristics that form the basis for our data assimilation implementation. Section \ref{sec:3} presents FARSITE data assimilation using EnSF, including the mathematical formulation, algorithmic details, and computational considerations necessary for real-time operation. Section \ref{sec: exp} evaluates the performance of our proposed method through comprehensive experiments. We begin with a simulation-based experiment in Section \ref{subsec:sim} to validate the method under controlled experimental conditions with known ground truth, followed by real wildfire applications in Section \ref{subsec:app} that demonstrate the practical effectiveness of our approach on historical fire events. Finally, Section \ref{sec:5} concludes the paper with a discussion of our findings, limitations of the current approach, and directions for future research in wildfire prediction and data assimilation.

\section{Data for FARSITE}
\label{sec:FARSITE}

Predictions using FARSITE hinge on a comprehensive suite of inputs: a landscape file defining topography, dynamic weather and wind data, and an initial fire perimeter. In this paper, we focus on the latest version of FARSITE \cite{mbedwardFARSITE}, and a key advancement in this version is its ability to forecast fire perimeter progression on an hourly basis, a significant improvement over the daily intervals of previous iterations. 

However, the frequent unavailability or incompleteness of required datasets requires meticulous data preparation and quality assurance. In what follows, we document the model's data requirements, identify gaps in our source datasets, and describe the pre-processing and substitution methods used to construct a complete and viable input for the simulation. To ensure model compatibility with this FARSITE version, all source data are converted to meet its specific unit requirements: English for weather and Metric for topography. 

\begin{table}[h!]
\centering
\renewcommand{\arraystretch}{1.2}
\begin{tabular}{lll}
\hline\hline
Data Type & Description & Unit \\
\hline
Elevation& Height above sea level & Meter \\
Aspect & Direction a slope faces & Degree \\
Slope & Steepness of terrain & Degree \\
Fuel Model & Classification of vegetation based on fire behavior & Class \\
Stand Height & Average height of trees & Meter \\
Canopy Cover & Percentage of ground covered by tree crowns & Percentage \\
Canopy Base Height &Height from ground to lowest tree branches & Meter \\
Canopy Bulk Density & Mass of available canopy fuel per volume & kg/m$^3$ \\
\hline
\end{tabular}
\caption{Description of the FARSITE landscape file components.}
\label{tab:farsite_landscape}
\end{table}

\textbf{Topographic data.}
The landscape file (.LCP) required for the FARSITE simulations is sourced from the Landscape Fire and Resource Management Planning Tools (LANDFIRE) program. The topographical and fuel data contained within this file are detailed in Table \ref{tab:farsite_landscape}.

The LANDFIRE program, which is jointly managed by the U.S. Department of Agriculture Forest Service and the U.S. Department of the Interior, provides nationally consistent spatial data that describes vegetation, wildland fuels, and fire regimes across the United States \cite{scott2005}.
For this study, we utilize the most current LANDFIRE Remap dataset to ensure the model inputs reflect recent landscape changes and disturbances. The data are processed at a 30-meter spatial resolution, a choice made to balance sufficient landscape detail with manageable computational demands for the FARSITE simulations.

The simulation domain for each wildfire event is defined as a 40 km $\times$ 40 km square area. This domain is centered on the geographic centroid of the final reported fire perimeter, ensuring the landscape fully encompasses the fire's ignition, spread, and eventual extent.

\vspace{6pt}
\textbf{Weather data.}
Weather data are from the Global Forecast System (GFS), accessed via Google Earth Engine (GEE) \cite{noaagfs0p25gee}. The GFS is a global numerical weather prediction model operated by the U.S. National Weather Service (NWS) that provides forecasts up to 16 days in advance. The specifications of the GFS data used in this work are detailed in Table \ref{tab:weather}.
\begin{table}[h!]
\centering
\renewcommand{\arraystretch}{1.2}
\begin{tabular}{lll}
\hline\hline
Data Type & Description & Unit \\
\hline
Cloud Cover & Total percentage of sky covered by clouds & Percentage \\
Precipitation &Cumulative amount of rain/snow & Inch \\
Relative Humidity &Air moisture content (2m above ground) & Percentage \\
Temperature &  Air temperature (2m above ground) & Fahrenheit \\
Wind Speed & Magnitude of wind & Mile per hour \\
Wind Direction  & Observed Wind Direction (clockwise from North) & Degree \\
\hline
\end{tabular}
\caption{Weather Data Obtained from the GFS Dataset.}
\label{tab:weather}
\end{table}

The GFS dataset provides observed data at six-hour intervals, though cloud cover and precipitation lack corresponding observed values. The dataset does include forecasted values for all parameters, which are generated based on the observed data at each time point. To address the gaps in observed data and to utilize the forecasted information, we calculate the median value across all available forecast horizons from the previous three days. This method applies to all forecasted values, including cloud cover and precipitation, where observed data would otherwise be expected but are not recorded in the dataset.

Furthermore, the GFS model does not provide wind speed and direction directly. Instead, it supplies the U-component (representing East-West wind) and V-component (representing North-South wind). These vector components are converted to wind speed and direction using standard trigonometric functions. 

\vspace{6pt}
\textbf{Fuel Moisture.}
Fuel moisture---the water content in vegetation---critically governs the fire ignition and spread. The FARSITE model applies these moisture values to characterize the fuel in each grid cell across the simulation domain, which presents a significant methodological challenge in obtaining precise, spatially explicit data for various fuel types.

In our evaluation of suitable data sources, we assessed the GridMET dataset \cite{abatzoglou2013}. Despite its high spatial resolution, GridMET exhibits considerable data gaps, particularly for the 10-hour, herbaceous, and woody fuel moisture categories required by the FARSITE model. These deficiencies render the dataset unsuitable for our research objectives.

Given these data limitations, we keep every fuel moisture variable constant for all experiments. For dead fuels, the moisture content for 1-hour, 10-hour, and 100-hour timelag categories is set to $6\%, 7\%$, and $8\%$, respectively. For live fuels, the moisture content for herbaceous (non-woody) plants is set to $60\%$ and for woody plants is set to $90\%$.

\vspace{6pt}
\textbf{Fire boundary data.}
Data assimilation requires the observed fire perimeters. The observational fire perimeter data are sourced from a dataset developed by the Center for Environmental Computing and Statistics (CECS), which contains VIIRS (Visible Infrared Imaging Radiometer Suite) satellite-derived boundaries for California wildfires (the final burned area $> 5$ km$^2$) from 2012 to 2020 \cite{chen2022}. In this dataset, fire perimeters are represented as either single polygon or multipolygon geometries. We note that the data for each fire does not capture the entire progression from ignition to extinguishment but rather a series of snapshots during its active period.

\vspace{6pt}
\textbf{Wildfire data.} The observational fire perimeter data are sourced from a dataset developed by the Center for Environmental Computing and Statistics (CECS), which contains VIIRS (Visible Infrared Imaging Radiometer Suite) satellite-derived boundaries for California wildfires (the final burned area $> 5$ km$^2$) from 2012 to 2020 \cite{chen2022}. 
To select suitable cases for our experiment, we apply several criteria. First, we focus on fires after July 1, 2016, to ensure compatibility with available GFS hourly weather data. Second, we exclusively select fires with single-polygon geometries. This process yields four wildfire use cases for our work, referred to by their Fire IDs: 1819, 2286, 3128, and 3179. The first two are considered small fires (final burned area $< 10$ km$^2$), while the latter two are classified as large.
\begin{table*}[h!]
\centering
\begin{tabular}{m{1.45in} l p{3.85in}}
\hline\hline
Parameter & Setting & Description \\
\hline
Timestep & 60 minutes & Maximum time conditions at a point are assumed constant for projecting the fire front position. \\
Distance Resolution & 30 meters & Controls the spacing of calculation points between time steps in the direction of fire spread. \\
Perimeter Resolution  & 60 meters & Controls the spacing of fire behavior calculation points along the simulated perimeter. \\
Minimum Ignition Vertex Distance & 15.0 & Minimum distance between ignition vertices. \\
Background Spotting Grid & 15 & Create a grid with the specified resolution and allow only one live ember per cell to minimize nearby spot fires. \\
Ember Spot& 0 & Probability of ember igniting a spot fire. Setting to 0 disables spot fires. \\
Spot Delay & 0 minutes & Simulates the delay before spot fires ignite and spread. \\
Minimum Spot Distance & 30 & Prevents live embers from igniting spot fires within this distance from the perimeter (if greater than zero). \\
Acceleration & 1 & When checked, the rate of spread increases gradually when conditions change. \\
Barriers & 0 & Controls whether small gaps in barriers are filled. \\
Foliar Moisture Content & 100 & Moisture content of the crown fuels; not related to the live fuel moisture of the surface fuels. \\
Crown Fire Calculation Method & Finney & Specifies the method, Finney (2004) or Scott/Reinhardt (2001), used to calculate crown fire behavior. \\
\hline
\end{tabular}
\caption{FARSITE Simulation Parameters}
\label{tab:farsite_parameters}
\end{table*}

Finally, 
Table~\ref{tab:farsite_parameters} details the FARSITE simulation parameters used in this work. These settings are held constant across all runs to ensure comparability, reproducibility, and minimal model variability for the subsequent data assimilation with EnSF.

\section{The Fire-EnSF Data Assimilation Framework }
\label{sec:3}
In this section, we present the data assimilation problem for FARSITE using EnSF. 
As discussed above, FARSITE represents the two-dimensional fire perimeter, at a discrete time step $k$ $(k=0,1,2,\cdots)$, by a polygon with several $n_k$ vertices. Let $h_k^{i}$  and $v_k^{i}$  respectively be the horizontal (east-west) and vertical (north-south) coordinates of the $i$-th vertex. We may define a vector $\bm{x}_k$ that defines the wildfire perimeter at time $k$,
\begin{equation}
    \bm{x}_k = \left( h_k^{1},v_k^{1},h_k^{2},v_k^{2} \,\cdots\, h_k^{n_k},v_k^{n_k}\right)^T \in \mathbb{R}^{2n_k}
\end{equation}
where $\bm{x}_0$ denotes the initial condition. It is noted that the dimension of the vector $\bm{x}_k$ changes over; also see Figure \ref{fig:dim}. 

During an active wildfire event, the fire perimeter can also be observed (with noise) by various approaches, such as remote sensing and UAVs. Using a similar approach, the observed fire perimeter can also be defined by a vector 
\begin{equation}
    \bm{y}_k = \left(\tilde{h}_k^{1},\tilde{v}_k^{1},\tilde{h}_k^{2},\tilde{v}_k^{2} \,\cdots\, \tilde{h}_k^{m_k},\tilde{v}_k^{m_k}\right)^T \in \mathbb{R}^{2m_k}
\end{equation}
where $m_k$ is the number of vertices of the observed fire perimeter, $\tilde{h}_k^{i}$  and $\tilde{v}_k^{i}$ are respectively the horizontal and vertical coordinates of the $i$-th vertex of the observed fire perimeter. Note that $m_k$ typically is not the same as $n_k$, and the value of $m_k$ also changes over time. 

The data assimilation problem for FARSITE is formulated as a classical dynamical model:
\begin{equation}
            \bm{x}_{k+1} = F(\bm{x}_k; \theta, \lambda_k), \quad
            \bm{y}_{k+1} = \bm{C} \bm{x}_{k+1} + \bm{e}_{k+1}. 
    \label{eq:state-space}
\end{equation}
Here, $\bm{x}_k$ is the state vector and $\bm{y}_k$ contains the observations. The evolution of the state vector $\bm{x}_k$ (with the change of its dimension)  is governed by a forward-prediction function $F(\cdot)$ within FARSITE. The function $F(\cdot)$ is an implicit, high-dimensional forward function with both static parameters $\theta$ (which encompasses static inputs like topography) and dynamic parameters $\lambda_k$ (such as weather conditions). The observation equation captures the relationship between the observation and the state. Following the work of \cite{SRIVAS2016897}, the observation equation in (\ref{eq:state-space}) takes a linear form with $\bm{C}$ being a matrix and $\bm{e}_{k} \sim N(0, \sigma_{obs}^2 \bm{I})$ captures the observation noise.

Next, we present the detailed descriptions of the steps (i.e., filtering recursions) involved in the data assimilation problem for the dynamic model (\ref{eq:state-space}). 

$\bullet$ [initialization] At any time $k$, draw $N$ sample vectors from the posterior distribution $P\left\{\bm{x}_k|\bm{y}_{1:k}\right \}$. We denote each sample by $\bm{x}_{k|k}^{i}\in \mathbb{R}^{2n_k}$, $i=1,2,\cdots,N$, which defines a fire perimeter polygon. If $k=0$, the samples are drawn from some prior distribution at time 0. Note that all sample vectors have the same dimension. 

$\bullet$ [one-step-ahead prediction for the states] Based on classical filtering recursions, this step involves generating the one-step-ahead predictive density for the states from
the filtered density 
\begin{equation} \label{eq:prediction_state}
P\{\bm{x}_{k+1}|\bm{y}_{1:k}\} = \int F(\bm{x}_k; \theta, \lambda_k)P\{\bm{x}_k|\bm{y}_{1:k}\}d\bm{x}_k.
\end{equation}
where $F(\cdot)$ is the function within FARSITE. 

In our implementation presented in the numerical examples, the prediction step initiates the filtering cycle by generating the prior ensemble for time $k+1$. This is achieved by advancing each member of the current filtered ensemble, $\{\bm{x}_{k|k}^{(i)}\}_{i=1}^{N}$, using the Euler-Maruyama discretization of the governing stochastic differential equation:
\begin{equation} \label{eq:prediction_step}
\bm{x}_{k+1|k}^{i} = \bm{x}_{k|k}^{i} + F(\bm{x}_k^{i}) \Delta k + \sqrt{\Delta k} \cdot \Sigma_{SDE} \cdot \bm{w}_k^{i}
\end{equation}
where each $\bm{w}_k^{i}$ is a random vector drawn from a standard normal distribution, $N(0, \bm{I})$, to model the process noise.

$\bullet$ [re-interpolation] As noted in \cite{SRIVAS2016897}, the observation $\bm{y}_{k+1}$ and the predicted state vectors $\bm{x}_{k+1}$, from the FARSITE model given $\bm{x}_{k}$, have mismatched dimensions (i.e., the polygons associated with different predicted fire perimeters have different number of vertices). This occurs because the polygons representing the $N$ predicted state samples, $\tilde{\bm{x}}_{k+1|k}^1, \dots, \tilde{\bm{x}}_{k+1|k}^N$, and the observed perimeter, $\bm{y}_{k+1}$, generally contain a different number of vertices, denoted $n_{k+1|k}^1, \dots, n_{k+1|k}^N$, and $m_{k+1}$, respectively. To resolve this dimensional inconsistency, we adopt the re-interpolation approach from \cite{SRIVAS2016897}. The first step is to identify the maximum vertex count across all polygons:
\begin{equation}
n_{k+1|k}=\max\left(n_{k+1|k}^1,n_{k+1|k}^2,\cdots,n_{k+1|k}^N, m_{k+1}\right)
\end{equation}
We then re-interpolate the vertices of the sampled state and observation vectors on the 2D spatial domain. This process yields a new set of re-interpolated state vectors, $\bm{x}_{k+1|k}^1, \bm{x}_{k+1|k}^2, \dots, \bm{x}_{k+1|k}^N$, and an observation vector, $\bm{y}_{k+1}$, which all share the same dimension. A detailed description of this procedure is provided in Section \ref{app:re-interpolation}.



$\bullet$ [state update (data assimilation)] The final step in the recursions involves drawing samples from the posterior filtering distribution
\begin{equation}
\label{eq:filtering}
P\{\bm{x}_{k+1}|\bm{y}_{1:k+1}\} = \frac{P\{\bm{y}_{k+1}|\bm{x}_{k+1}\}P\{\bm{x}_{k+1}|\bm{y}_{k}\}}{P\{\bm{y}_{k+1}|\bm{y}_{1:k}\}}.
\end{equation}
where the one-step-ahead predictive density of the observations above is given by 
\begin{equation}
\label{eq:prediction_observation}
P\{\bm{y}_{k+1}|\bm{y}_{1:k}\} = \int P\{\bm{y}_{k+1}|\bm{x}_{k+1}\}P\{\bm{x}_{k+1}|\bm{y}_{1:k}\} \, d\bm{x}_{k+1}.
\end{equation}

The state filtering step above usually poses a significant computational challenge for high-dimensional problems. The existing EnKF hinges on the idea of representing the state distribution by a sample of ``ensemble'' \cite{katzfuss2016understanding}. The recent development of the new EnSF shows that such an ensemble of samples can be efficiently and accurately generated from a diffusion process \cite{BAO2024}. 

\vspace{6pt}
The EnSF captures the evolution of the filtering density based on a score-based diffusion model. It learns a stochastic transport map that transforms a simple distribution (e.g., a standard Gaussian) into the complex posterior distribution of the system state. This is achieved via a pair of forward and reverse-time Stochastic Differential Equations (SDEs) defined over a pseudo-time domain $\tau \in [0, 1]$. 

Following the work of \cite{BAO2024}, the forward process, which adds noise to the state $\bm{x}_k$ over the pseudo-time domain, is given by an SDE:
\begin{equation}\label{eq:forward_sde}
    d\bm{x}_{k, \tau} =  \bm{x}_{k,\tau}d\log \alpha_{\tau} + \left( \frac{d\beta_\tau^2}{d\tau}-2\frac{d \log\alpha_\tau}{d\tau}\beta_\tau^2\right)^{1/2}d\bm{w}
\end{equation}
where $\alpha_\tau = 1 - \tau(1-\epsilon_\alpha)$ and $\beta^2_\tau = \epsilon_\beta + \tau(1-\epsilon_\beta)$ define the mean and variance of the reverse diffusion process with positive (tuning) hyperparameters $\epsilon_\alpha$ and $\epsilon_\beta$ for numerical stability, 
and $\bm{w}$ is a standard Wiener process.

The reverse-time SDE, which transforms samples from the simple noise distribution at $\tau=1$ back to samples from the original data distribution at $\tau=0$, is given by 
\begin{equation}
    d\bm{x}_{k} = \left[ \frac{d\log \alpha_\tau}{d\tau}\bm{x}_{k,\tau} - \left(\frac{d\beta_\tau^2}{d\tau}-2\frac{d \log\alpha_\tau}{d\tau}\beta_\tau^2 \right)\nabla_{\bm{x}}\log P_\tau(\bm{x}) \right]d\tau + \left( \frac{d\beta_\tau^2}{d\tau}-2\frac{d \log\alpha_\tau}{d\tau}\beta_\tau^2\right)^{1/2}d\bar{\bm{w}}
    \label{eq:reverse_sde}
\end{equation}
where $\bar{\bm{w}}$ is a reverse-time Wiener process and $\nabla_x \log P_\tau(x)$ is known as the score function that needs to be determined.

The following score function has been proposed to sample the $\bm{x}_{k+1}$ from the filtering distribution $P\{\bm{x}_{k+1}|\bm{y}_{1:k+1}\}$ in (\ref{eq:filtering}):
\begin{equation} \label{eq:score_post}
    \nabla_{\bm{x}} \log P\{\bm{x}_{k+1,\tau} | \bm{y}_{1:k+1}\} = \nabla_{\bm{x}} \log P\{\bm{x}_{k+1,\tau} | \bm{y}_{1:k}\} + h(\tau)\nabla_{\bm{x}} \log P\{\bm{y}_{k+1} | \bm{x}_{k+1,\tau}\}
\end{equation}
where $h(\tau) = 1 - \tau$ is a monotonically decreasing damping function that anneals the influence of the new observation. Here, the first term is known as the prior score function, while the second term is the likelihood. For this reason, the score function in (\ref{eq:score_post}) is known as the posterior score function, which is updated from the prior score function given the likelihood. 

In particular, the prior score function for each sample $\bm{x}^{i}_{k+1, \tau}$ is approximated as follows, given the one-step-ahead predicted states (after the interpolation of vertices of the fire perimeters above)
\begin{equation}
    \nabla_{\bm{x}} \log P\{\bm{x}_{k+1,\tau}^{i} | \bm{y}_{1:k}\} \approx \nabla_{\bm{x}} \log P\{\bm{x}_{k+1,\tau}^{i} | \bm{x}_{k+1|k}^{i}\} = - \frac{\bm{x}_{k+1,\tau}^{i} - \alpha_\tau \bm{x}_{k+1, 0}^{i}}{\beta^2_\tau}
    \label{eq:prior_score_simplified}
\end{equation}
and the likelihood  is derived from the observation equation in (\ref{eq:state-space}):
\begin{equation}
    \nabla_{\bm{x}} \log P\{\bm{y}_{k+1} | \bm{x}_{k+1,\tau}\} = \bm{C} \frac{(\bm{y}_{k+1}-\bm{C}\bm{x}_{k+1, \tau} )}{\sigma_{obs}^2}. 
    \label{eq:likelihood_score}
\end{equation}

\begin{figure}[h!]
    \centering
\includegraphics[width=\linewidth]{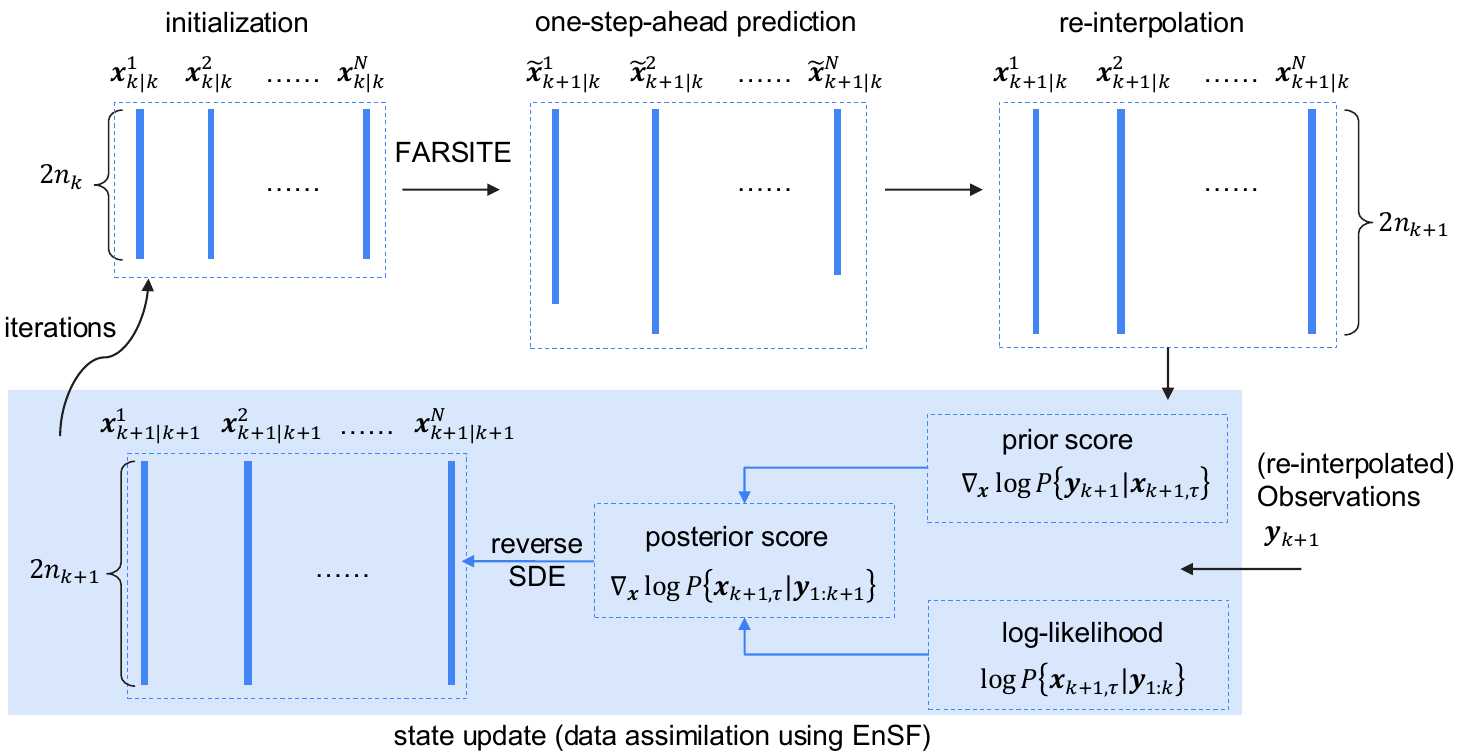}
    \vspace{-12pt}
    \caption{Data assimilation for FARSITE using the EnSF}
    \label{fig:process}
\end{figure}

Figure \ref{fig:process} provides a summary of the data assimilation steps described above. 
\vspace{-2pt}

\section{Additional Implementation Details}
In this section, we provide details on the re-interpolation step (of the vertices of fire perimeter polygons) as well as the polygon vertex normalization. 

\subsection{Vertices Re-interpolation}
\label{app:re-interpolation}

The vertices re-interpolation step is required to address the issue of the time-varying dimension of fire perimeter polygons. 
Vertices re-interpolation samples a given polygon, represented by an ordered list of $M$ vertices, to a new polygon with a specified number of $M'$ vertices. The new vertices are distributed uniformly along the perimeter of the original polygon, preserving its overall shape.
Let the input polygon be defined by a sequence of vertices $P = (\mathbf{v}_1, \mathbf{v}_2, \dots, \mathbf{v}_M)$, where each vertex $\mathbf{v}_i \in \mathbb{R}^2$ is given by its coordinates $(v_i^x, v_i^y)$. The objective is to generate a new sequence of vertices $P' = (\mathbf{v}'_1, \mathbf{v}'_2, \dots, \mathbf{v}'_{M'})$.

To accurately measure the perimeter, the polygon is first treated as a closed loop. This is achieved by creating a closed sequence of vertices where the first vertex is appended to the end, forming a path from $\mathbf{v}_M$ back to $\mathbf{v}_1$. Let this closed sequence be denoted by $P_c = (\mathbf{v}_1, \mathbf{v}_2, \dots, \mathbf{v}_M, \mathbf{v}_{M+1})$, where $\mathbf{v}_{M+1} = \mathbf{v}_1$.

The Euclidean distance between each pair of consecutive vertices $(\mathbf{v}_i, \mathbf{v}_{i+1})$ for $i=1, \dots, M$ is calculated to determine the length of each segment, $s_i$:
\begin{equation}
    s_i = \|\mathbf{v}_{i+1} - \mathbf{v}_i\|_2 = \sqrt{(v_{i+1}^x - v^x_i)^2 + (v_{i+1}^y - v^y_i)^2}.
\end{equation}

Next, a cumulative distance array, $\{D_i\}_{i=1}^{M+1}$, is computed. This array stores the total distance from the starting vertex $\mathbf{v}_1$ to every other vertex along the perimeter. It is defined as $D_k = \sum_{i=1}^{k-1} s_i$ for $k > 1$, with the initial condition $D_1 = 0$. The total perimeter of the polygon, $L$, is the final value in the cumulative sum:
\begin{equation}
    L = \sum_{i=1}^{M} s_i = D_{M+1}
\end{equation}

With the total perimeter $L$ known, a set of $M'$ target distances, $\{d_j\}_{j=1}^{M'}$, is generated. These distances are uniformly spaced along the perimeter from the starting point to the end. This is achieved by creating a linear space:
\begin{equation}
    d_j = L \times \frac{j-1}{N-1}, \quad \text{for } j = 1, 2, \dots, M'
\end{equation}
Note that, $d_1 = 0$ and $d_{M'} = L$.

For each target distance $d_j$, the corresponding new vertex $\mathbf{v}'_j$ is found via linear interpolation. First, we identify the line segment $[\mathbf{v}_k, \mathbf{v}_{k+1}]$ that contains the point at distance $d_j$. This is the segment for which the cumulative distances satisfy the condition $D_k \le d_j \le D_{k+1}$.
Given the interpolation factor, $\alpha \in [0, 1]$, which represents the fractional distance of $d_j$ along the segment $s_k$,
\begin{equation}
    \alpha = \frac{d_j - D_k}{D_{k+1} - D_k} = \frac{d_j - D_k}{s_k},
\end{equation}
the new vertex $\mathbf{v}'_j = (x'_j, y'_j)$ is then calculated by linearly interpolating between the vertices $\mathbf{v}_k$ and $\mathbf{v}_{k+1}$:
\begin{equation}
    \mathbf{v}'_j = (1 - \alpha)\mathbf{v}_k + \alpha\mathbf{v}_{k+1}.
\end{equation}

This operation is performed independently on the horizontal and vertical coordinates. The resulting collection of vertices, $P' = (\mathbf{v}'_1, \mathbf{v}'_2, \dots, \mathbf{v}'_{M'})$, forms the resampled polygon with uniformly spaced vertices.

\subsection{Polygon Vertex Normalization}

Polygon vertex normalization normalizes the vertex representation of a polygon, which is assumed to be defined by a pre-sequenced list of its vertices. The normalization process enforces two conditions: (i) the vertices are ordered in a clockwise direction, and (ii) the sequence begins with a canonical starting vertex, specifically the vertex closest to the positive \textit{x}-axis.

Similar to Section \ref{app:re-interpolation}, let the input polygon be given by a sequence of $M$ vertices $P = (\mathbf{v}_1, \mathbf{v}_2, \dots, \mathbf{v}_M)$. 
The winding order of the polygon's vertices is determined using the surveyor's formula, also known as the \textit{shoelace} formula, which calculates the signed area of a simple polygon. The signed area, $A$, is given by:
\begin{equation}
    A = \frac{1}{2} \sum_{i=1}^{M} (v^x_i v^y_{i+1} - v^x_{i+1} v^y_i)
\end{equation}
where $(v^x_{M+1}, v^y_{M+1})$ are the same as $(v^x_1, v^yy_1)$. By convention, a positive area ($A > 0$) corresponds to a counter-clockwise (CCW) winding order, while a negative area corresponds to a clockwise (CW) order.
To enforce a consistent CW ordering, the sign of the calculated area is checked. If $A > 0$, indicating a CCW order, the sequence of vertices is reversed. Let the resulting sequence, which is guaranteed to be in CW order, be denoted by $P_{CW} = (\mathbf{u}_1, \mathbf{u}_2, \dots, \mathbf{u}_{M})$.

The next step is to re-order the sequence so that it begins from a uniquely defined starting point. The chosen canonical start is the vertex that forms the smallest angle with the positive x-axis.
For each vertex $\mathbf{u}_i = (u^x_i, u^y_i)$ in the clockwise-ordered sequence $P_{CW}$, the angle $\theta_i$ with respect to the origin is calculated using the two-argument arctangent function
\begin{equation}
    \theta_i = \operatorname{atan2}(u^y_i, u^x_i)
\end{equation}
which maps the coordinates to the full range of angles $(-\pi, \pi]$.

To find the vertex closest to the positive \textit{x}-axis, we identify the index $k$ that corresponds to the minimum absolute angle:
\begin{equation}
    k = \arg\min_{i \in \{1, \dots, M\}} |\theta_i|
\end{equation}
and the vertex $\mathbf{u}_k$ is then selected as the new starting vertex of the normalized polygon.

The final normalized vertex sequence, $P'$, is obtained by performing a cyclic shift on the clockwise-ordered list $P_{CW}$ such that the vertex $\mathbf{u}_k$ becomes the first element. The new sequence is given by
\begin{equation}
    P' = (\mathbf{u}_k, \mathbf{u}_{k+1}, \dots, \mathbf{u}_N, \mathbf{u}_1, \dots, \mathbf{u}_{k-1}).
\end{equation}

\section{Numerical Experiments}
\label{sec: exp}

Numerical investigations are performed to demonstrate the performance of data assimilation for FARSITE using EnSF. 
The first example is based on simulated data while the second involves the real data from past wildfire episodes. Comparison studies, in terms of accuracy and computational time, are conducted between the approach described in this paper and the existing ones. 

\subsection{Haversine Distances}

Before presenting the numerical examples, this section details the methods used to compute the quantitative error between two sets of corresponding geographic coordinates. The primary distance metric employed is the Haversine distance, which is then used to calculate the Root Mean Square Error (RMSE).
The Haversine formula is used to calculate the great-circle distance between two points on a sphere, given their longitudes and latitudes. This is particularly well-suited for determining the separation between geographic coordinates.

Let there be two corresponding sets of $p$ points, where the coordinates for each point are given in degrees as $(\theta, \phi)$, representing longitude and latitude, respectively. For any pair of corresponding points $(\theta_1, \phi_1)$ and $(\theta_2, \phi_2)$, the first step is to convert these coordinates to radians.

The differences in longitude and latitude are then calculated:
\begin{equation}
\Delta\theta = \theta_{2,rad} - \theta_{1,rad}, \quad \Delta\phi = \phi_{2,rad} - \phi_{1,rad}
\end{equation}

The core of the Haversine formula calculates an intermediate value, 
\begin{equation}
    a = \sin^2\left(\frac{\Delta\phi}{2}\right) + \cos(\phi_{1,rad}) \cos(\phi_{2,rad}) \sin^2\left(\frac{\Delta\theta}{2}\right)
\end{equation}
and a central angle, $c$, between the two points:
\begin{equation}
    c = 2 \cdot \operatorname{atan2}(\sqrt{a}, \sqrt{1 - a}).
\end{equation}

The final distance, $d$, is computed by multiplying the central angle by the Earth's mean radius, $R$ (approximately 6371 km):
\begin{equation}
    d = R \cdot c.
\end{equation}

To aggregate the individual Haversine distances into a single error metric, the Root Mean Square Error is computed. The RMSE provides a measure of the average magnitude of the error between the two sets of points.
Given an array of $p$ Haversine distances, $\{d_1, d_2, \dots, d_p\}$, calculated between each corresponding pair of points, the RMSE is defined as:
\begin{equation}
    \text{RMSE} = \left(\frac{1}{p} \sum_{i=1}^{p} d_i^2\right)^{1/2}
\end{equation}
which is used as the error metric in the following numerical examples. 

\subsection{Example I: A Simulation-Based Experiment}
\label{subsec:sim}
\textbf{Experiment Setup.}
The first example is based on a simulation dataset. The goal of this example is to demonstrate how the EnSF-based data assimilation works and how it is compared with the more conventional EnKF-based approach. Under a well controlled environment, the true system state is known, which allows for the assessment of the accuracy of the estimated state. 

Consider a simple dynamical model in the form of (\ref{eq:state-space}) as follows 
\begin{equation}
\begin{split}
    \bm{x}_{k+1} & = 2\bm{x}_k \\
    \bm{y}_{k+1} & = 0.25 \bm{x}_{k+1} + \bm{e}_{k+1}
\end{split}
\label{eq:sim_set}
\end{equation}
where $\bm{e}_{k+1} \sim N(0, 0.01\bm{I})$. 

To address the computational challenge of changing dimensions over time in real applications, we design our simulation with a fixed 2500-dimensional state space using 100 ensemble members. The high dimensionality of the problem significantly challenges the accuracy and computational efficiency of the data assimilation approaches. We employ a discrete time increment of 0.05 over 30 data assimilation (i.e., filtering) steps. This fixed-dimension setup eliminates the need for vertex matching and dimension changes at each step, while providing a computationally demanding test case for both data assimilation methods.

We evaluate the performance of data assimilation using both the EnKF and EnSF using the Root Mean Squared Error. For the EnSF, the reverse diffusion process is discretized into 100 iterations with $\Sigma_{SDE} = 0.5$. When computing the RMSE, the state coordinates are interpreted as latitude-longitude pairs, which provides a geographically meaningful error measure.



\begin{figure}[h!]
\centering 
\begin{subfigure}[b]{0.496\textwidth} 
    \includegraphics[width=\textwidth]{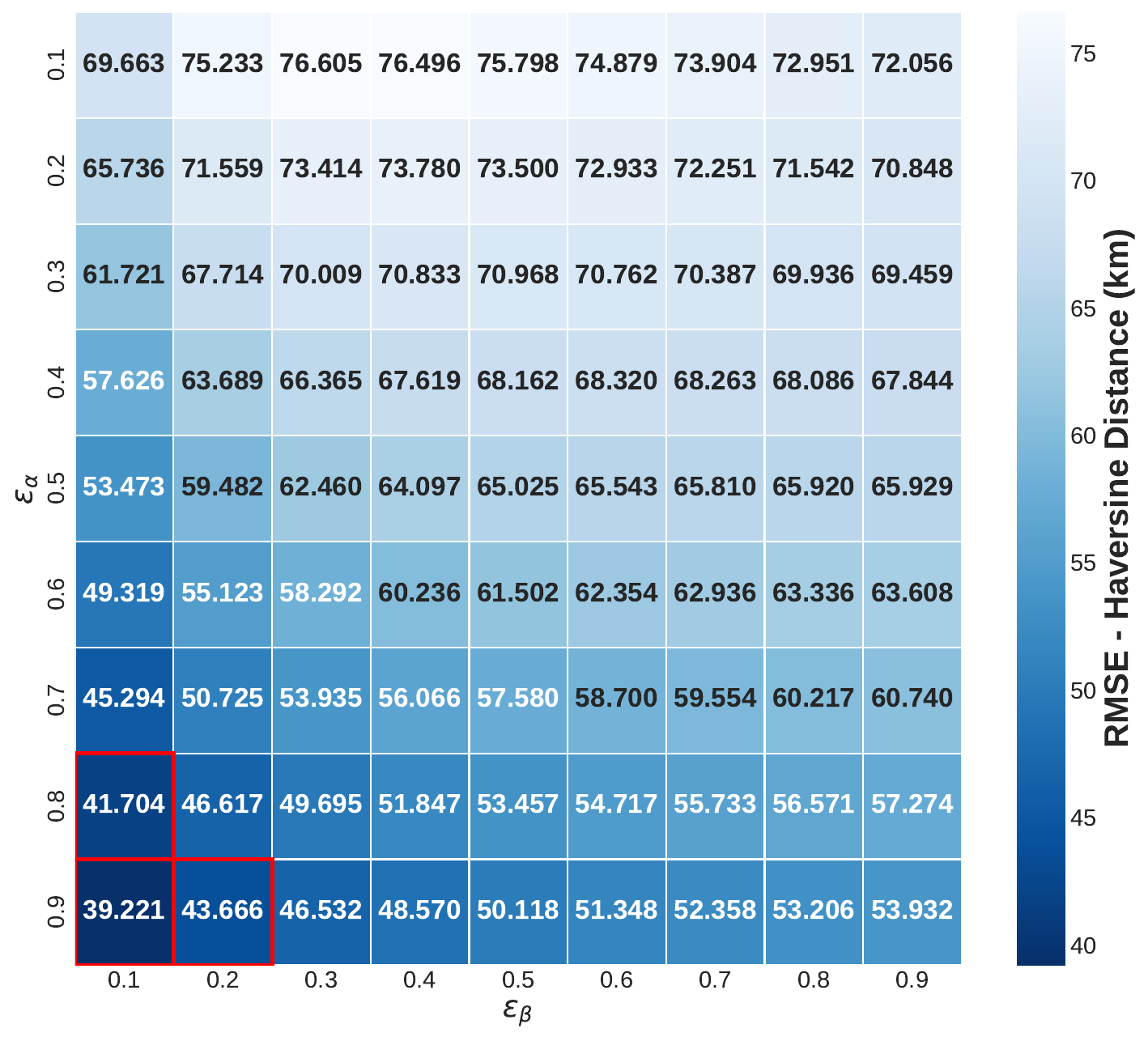} 
\end{subfigure}
\hfill 
\begin{subfigure}[b]{0.496\textwidth}
    \includegraphics[width=\textwidth]{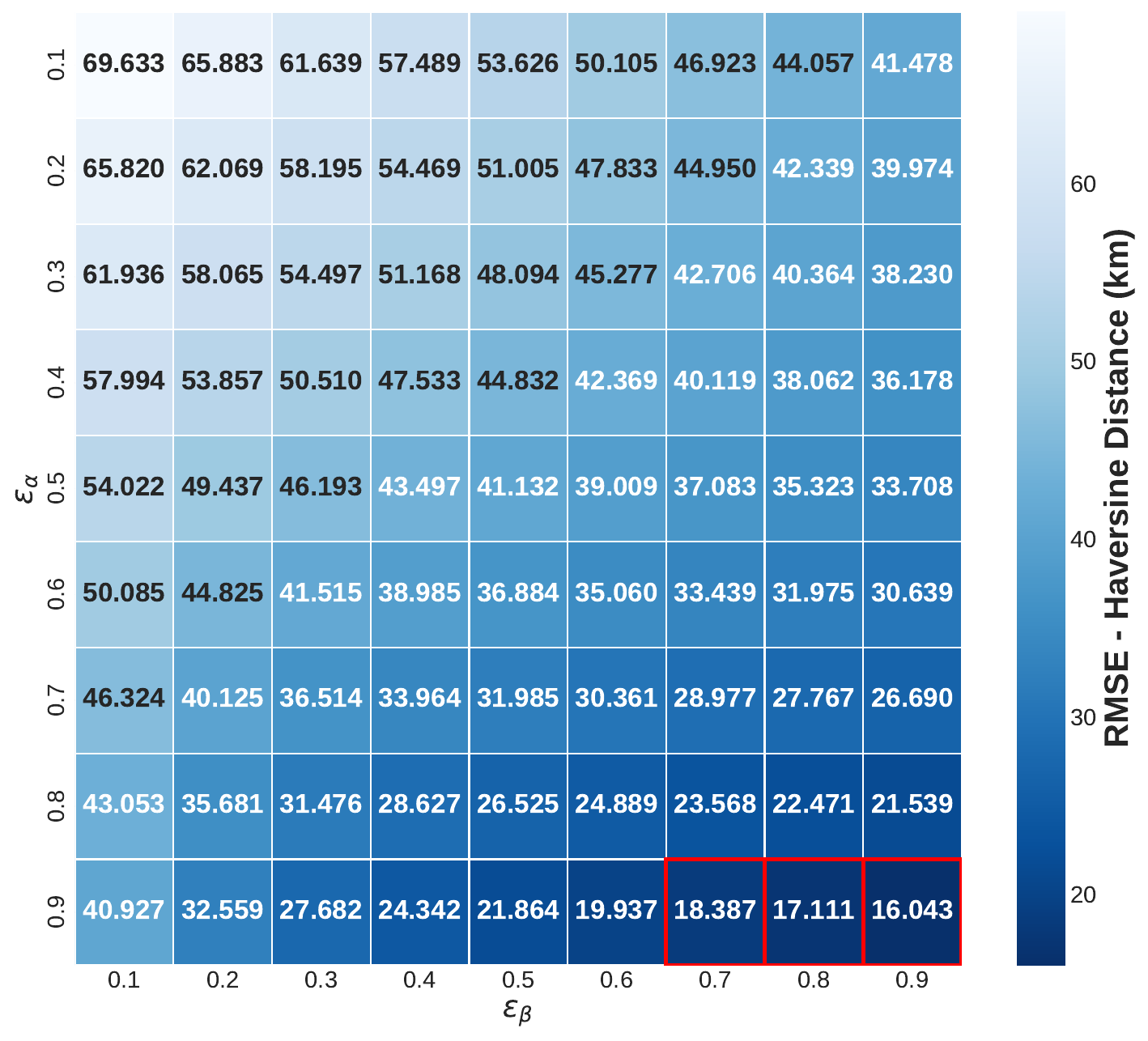}
\end{subfigure}
\caption{Hyperparameter tuning results for EnSF with the best 3 options circled with a red square. Left: RMSE between ground truth and estimation; Right: RMSE between observation and estimation.}
\label{fig: sim_tune}
\end{figure}

\textbf{Experiment Results.} We begin with coarse-tuning of the hyperparameters by incrementing both $\epsilon_\alpha$ and $\epsilon_\beta$ between 0 and 1 with steps of 0.1. Figure \ref{fig: sim_tune} presents the average RMSE between the ground truth and estimated states (left panel), and that between the observed and estimated states (right panel), over all filtering steps. 
The left panel of Figure \ref{fig: sim_tune} indicates that the optimal region is found near the lower-left corner, where $\epsilon_\alpha$ approaches 1 and $\epsilon_\beta$ approaches 0. In general, the RMSE increases diagonally towards the upper-right corner. 
In contrast, the right panel of Figure \ref{fig: sim_tune} shows that the optimal region is found near the upper-right corner, where both hyperparameters approach 1, with the RMSE increasing diagonally towards the lower-left corner. 

\begin{figure}[h!]
\centering
\includegraphics[width=0.7\linewidth]{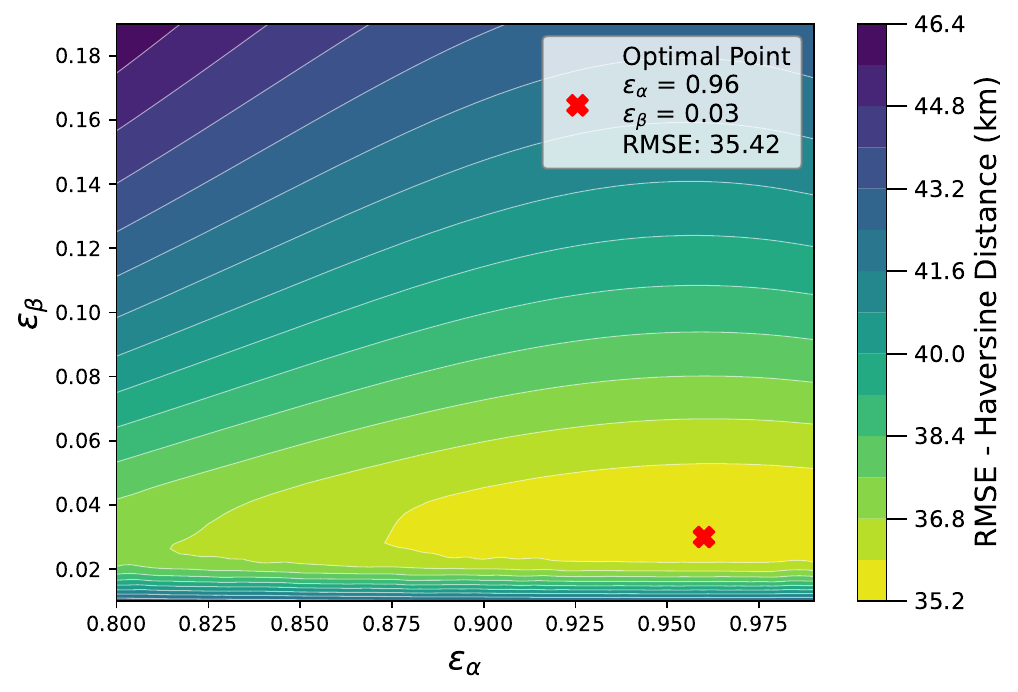}
    \captionof{figure}{Hyperparameter tuning for EnSF showing the RMSE between the estimated states and the ground truth.}
    \label{fig:sim_ftune}
\end{figure}

\begin{figure}[h!]
\centering
\includegraphics[width=0.7\linewidth]{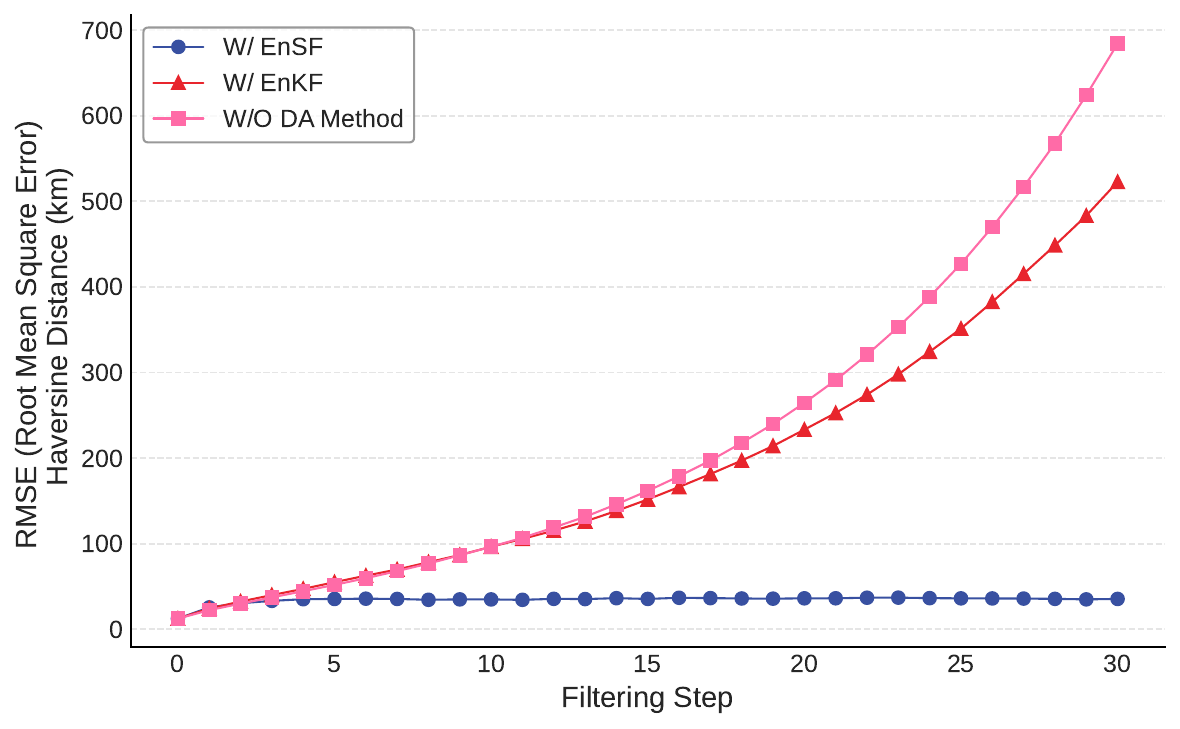}
    \captionof{figure}{RMSE comparison between the estimated states and the ground truth.}
    \label{fig:sim_com}
\end{figure}

Based on the observations above, we proceed with fine-tuning $\epsilon_\alpha$ between 0.8 and 1.0, and $\epsilon_\beta$ between 0.0 and 0.2, with increments of 0.01, and find the optimal hyperparameters $\epsilon_\alpha = 0.96$ and $\epsilon_\beta = 0.03$ (see Figure \ref{fig:sim_ftune}).  Based on the fine-tuned parameters, Figure \ref{fig:sim_com} presents a quantitative comparison of the RMSE between the estimated and ground truth values. It is seen that data assimilation using the EnSF provides a stable forecasting performance with an RMSE significantly lower than that of the original dynamic system without data assimilation. While the EnKF shows only a slight reduction in error compared to the baseline system, its performance is substantially inferior to that of the EnSF.

\textbf{Computational Efficiency.} Computational efficiency is evaluated using an Intel Xeon CPU and an NVIDIA RTX A6000 GPU. The investigation reveals significant advantages for EnSF, with a mean execution time of $0.672 \pm 0.012$ seconds compared to $3.634 \pm 0.013$ seconds for EnKF. This represents approximately a 5.4-fold speed improvement, with the small standard deviations indicating highly consistent and reliable measurements across test runs. The dramatic performance difference suggests that EnSF offers superior algorithmic efficiency on the GPU architecture, which makes it particularly valuable for later real-time applications that require numerous repeated calculations where computational speed is critical.


In summary, the findings from the simulation-based experiment show the potential of using the EnSF for data assimilation with FARSITE. The method not only achieves better accuracy in state estimation but also delivers an 82\% reduction in computational time with significantly greater stability, needed for operational wildfire monitoring systems. 

\subsection{Example II: Application to Real Wildfire Data}
\label{subsec:app}
\textbf{Data and Experiment Setup.} In this section, we investigate the data assimilation for FARSITE using EnSF. 
The observational fire perimeter data are sourced from a dataset developed by the Center for Environmental Computing and Statistics (CECS), which contains VIIRS (Visible Infrared Imaging Radiometer Suite) satellite-derived boundaries for California wildfires (the final burned area $> 5$ km$^2$) from 2012 to 2020 \cite{chen2022}. In this dataset, fire perimeters are represented as either single polygon or multipolygon geometries. We note that the data for each fire does not capture the entire progression from ignition to extinguishment but rather a series of snapshots during its active period.

To select suitable cases for our experiment, we focus on fires after July 1, 2016, to ensure compatibility with available GFS hourly weather data. We also exclusively select fires with single-polygon geometries. This process yields four wildfire use cases, referred to by their fire IDs: 1819, 2286, 3128, and 3179. The first two are small fires (final burned area $< 10$ km$^2$), while the latter two are considered as large fires. 

The dynamical model (\ref{eq:state-space}) is adopted with $\bm{C}=\bm{I}$ and $\bm{e}_{k+1} \sim N(0, 0.25 \bm{I})$. 
In particular, the forecast intervals are set to 0.5 days (half-day intervals). For the reverse diffusion process involved in the EnSF, we include 200 iterations with $\Sigma_{SDE} = 1$. 
\begin{figure}[h!]
\centering 
\begin{subfigure}[b]{0.6\textwidth} 
\includegraphics[width=\textwidth]{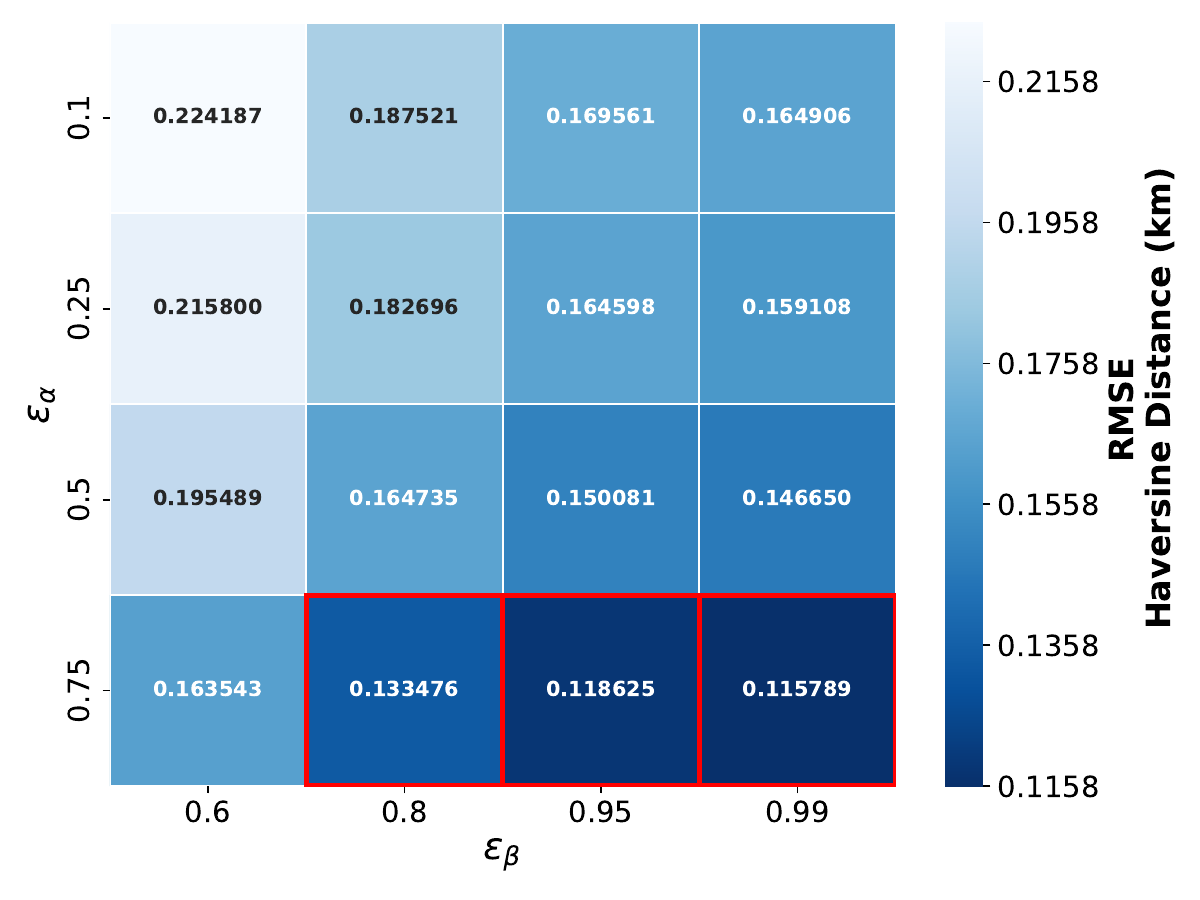} 
\end{subfigure}
\vfill 
\begin{subfigure}[b]{0.6\textwidth}
\includegraphics[width=\textwidth]{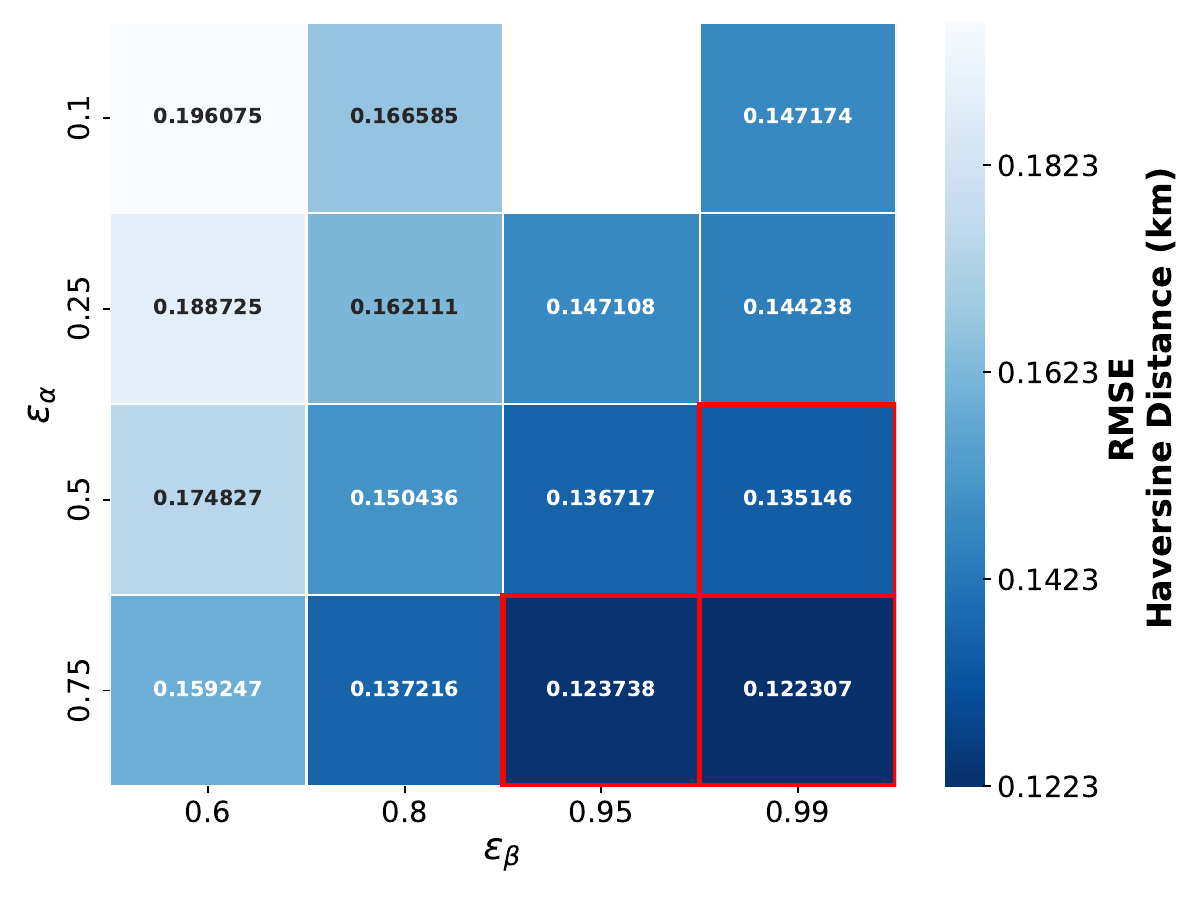}
\end{subfigure}
\caption{Hyperparameter tuning results showing RMSE values for different combinations of $\epsilon_\alpha$ and $\epsilon_\beta$ parameters for fires 1819 (top) and 2286 (bottom) between observation and estimation.}
\label{fig:app_tune}
\end{figure}
Figure \ref{fig:app_tune} shows the hyperparameter tuning results using four values for each parameter ($\epsilon_\alpha$ and $\epsilon_\beta$) in the analysis between the observed and estimated fire perimeters. The results reveal that larger values of both $\epsilon_\alpha$ and $\epsilon_\beta$ generally produce smaller RMSE values, while RMSE increases diagonally toward the upper-right corner. This finding is consistent with our previous simulation experiments, suggesting that similar results may be expected for state target estimation. Based on these results, we selected the optimal hyperparameter values of $\epsilon_\alpha = 0.96$ and $\epsilon_\beta = 0.03$ for comparison with the EnKF results.

Note that FARSITE is sometimes unstable for the version we used. This issue is demonstrated in fire 2286, where $\epsilon_\alpha = 0.1$ and $\epsilon_\beta = 0.95$, resulting in a computationally prohibitive model that requires approximately two weeks to process even small fires, unlike in other tests. Although the model would eventually converge, we observed significant performance degradation at the second filtering step. Given computational constraints, we decided not to pursue further since the pattern of the error has been determined.

\begin{figure}[!ht]
    \centering
    \includegraphics[width=0.8\linewidth]{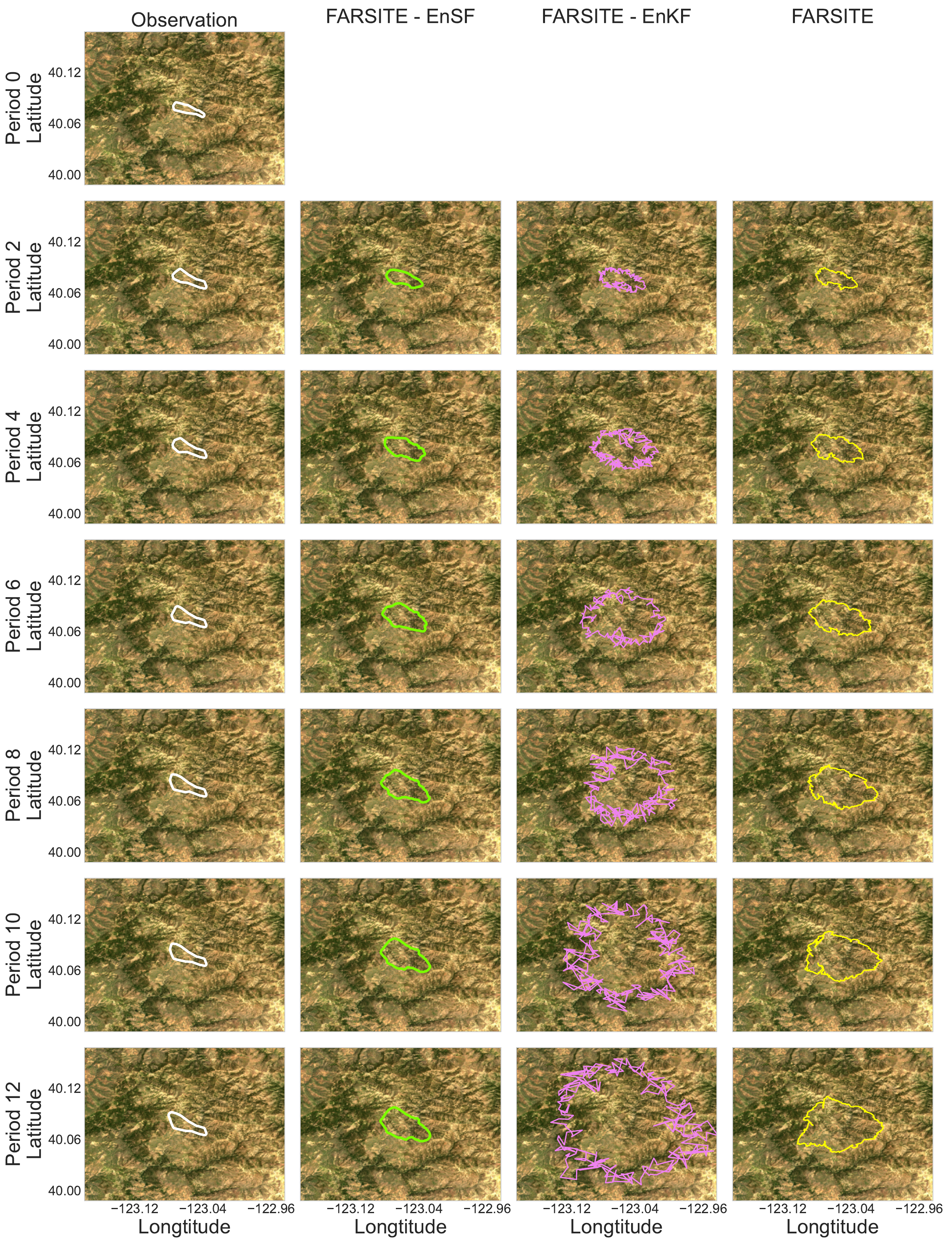}
    \caption{Comparison of fire perimeter estimations using (0.5 days forecasting interval) FARSITE and EnSF methods for fires 1819. Background imagery is sourced from LANDSAT.}
    \label{fig:1819}
\end{figure}

\begin{figure}[!ht]
    \centering
    \includegraphics[width=0.8\linewidth]{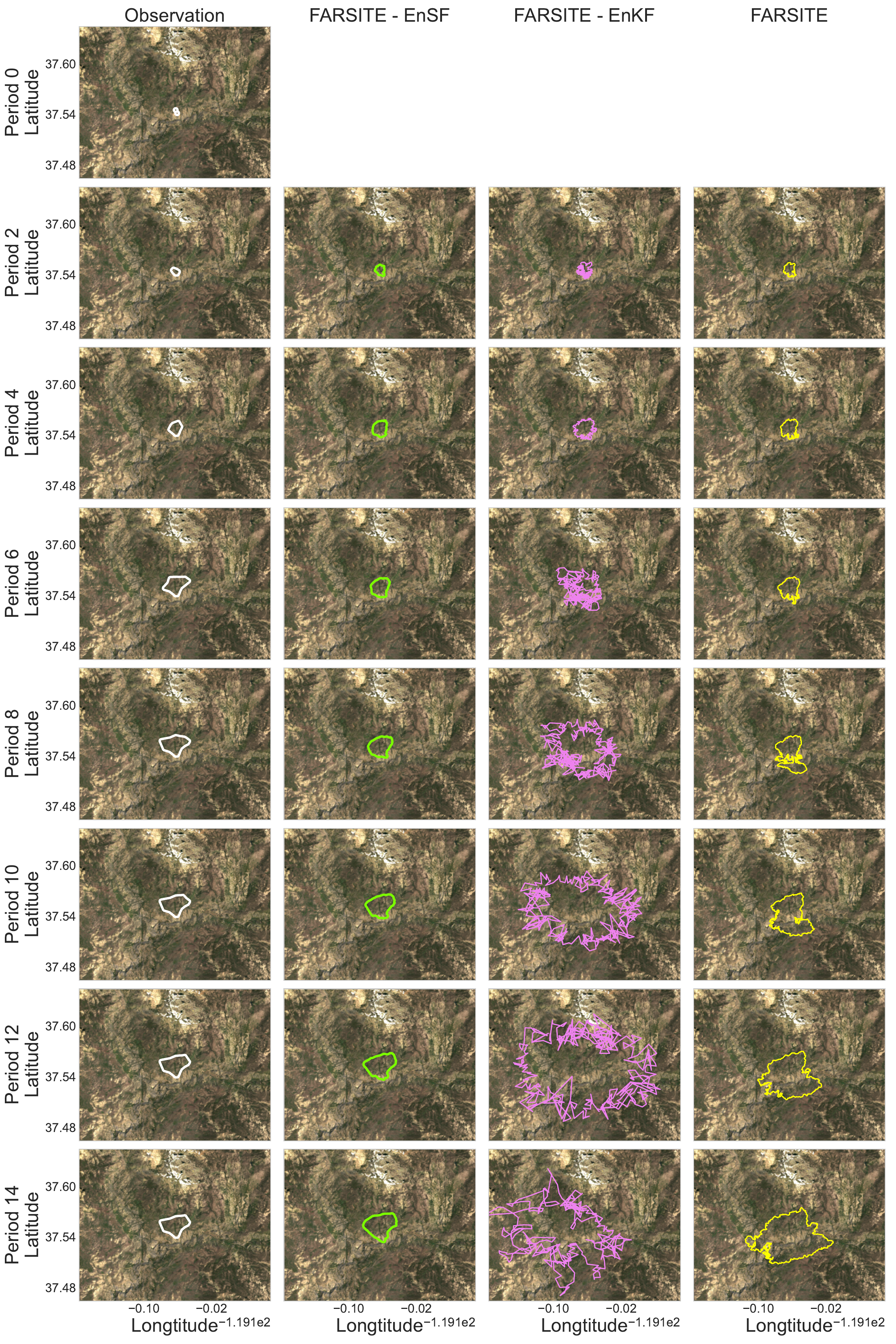}
    \caption{Comparison of fire perimeter estimations using (0.5 days forecasting interval) FARSITE and EnSF methods for fires 2286. Background imagery is sourced from LANDSAT.}
    \label{fig:2286}
\end{figure}

\textbf{Experiment Results.}
Figure \ref{fig:1819} and \ref{fig:2286} illustrates the fire perimeter estimations using FARSITE with EnSF, EnKF, and the standalone FARSITE for fires 1819 and 2286. While FARSITE's predictions deviate significantly from observations, EnSF successfully reduces the error between observations and FARSITE's original output, providing a more accurate estimation of the potential ground truth. In contrast, EnKF fails to eliminate estimation errors and instead produces estimates with both the largest area coverage and irregular polygon geometry.

The irregular polygon formation in EnKF results from its poor error elimination performance. The polygon vertex normalization implementation assumes that the original polygon is a perfect Hamiltonian circuit. When EnKF cannot adequately reduce errors, it breaks this circuit assumption. Since FARSITE is a level-set-based fire spread model that computes vertex positions sequentially, incorrect previous estimations propagate errors to subsequent forecasts.

This performance difference is also reflected in computational efficiency. Data assimilation using EnSF completes processing for such fires within an hour, while data assimilation using EnKF, which provides improper input at certain filtering steps and increases forecasting difficulty for FARSITE, requires at least two days with high risk of computational failure---similar to what occurred with fire 2286 when we implement EnSF with $\epsilon_\alpha = 0.1$ and $\epsilon_\beta = 0.95$, where one of the ensembles has been depreciated in fire 2286 at the last filtering step. Thus, the ensemble is not used for the last filtering step, which means we estimate the final boundary with only 29 ensemble members. This is also the reason that the area of the final boundary is much less than that in period 12.

\begin{figure}[h!]
    \centering
    \includegraphics[width=0.65\linewidth]{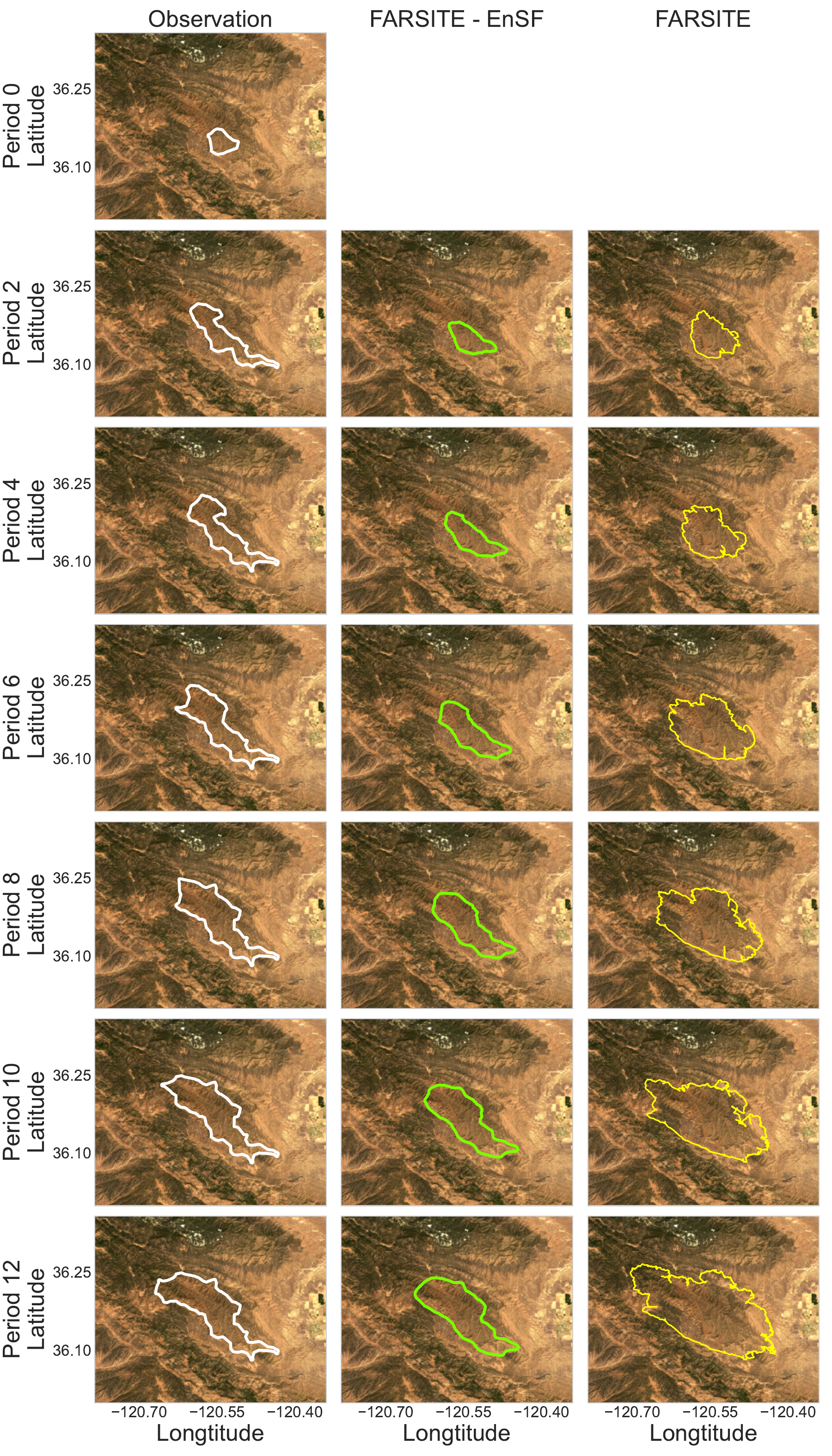}
    \caption{Comparison of fire perimeter estimations using (0.5 days forecasting interval) FARSITE and EnSF methods for fires 3128. Background imagery is sourced from LANDSAT.}
    \label{fig:3128}
\end{figure}
\begin{figure}[h!]
    \centering
    \includegraphics[width=0.7\linewidth]{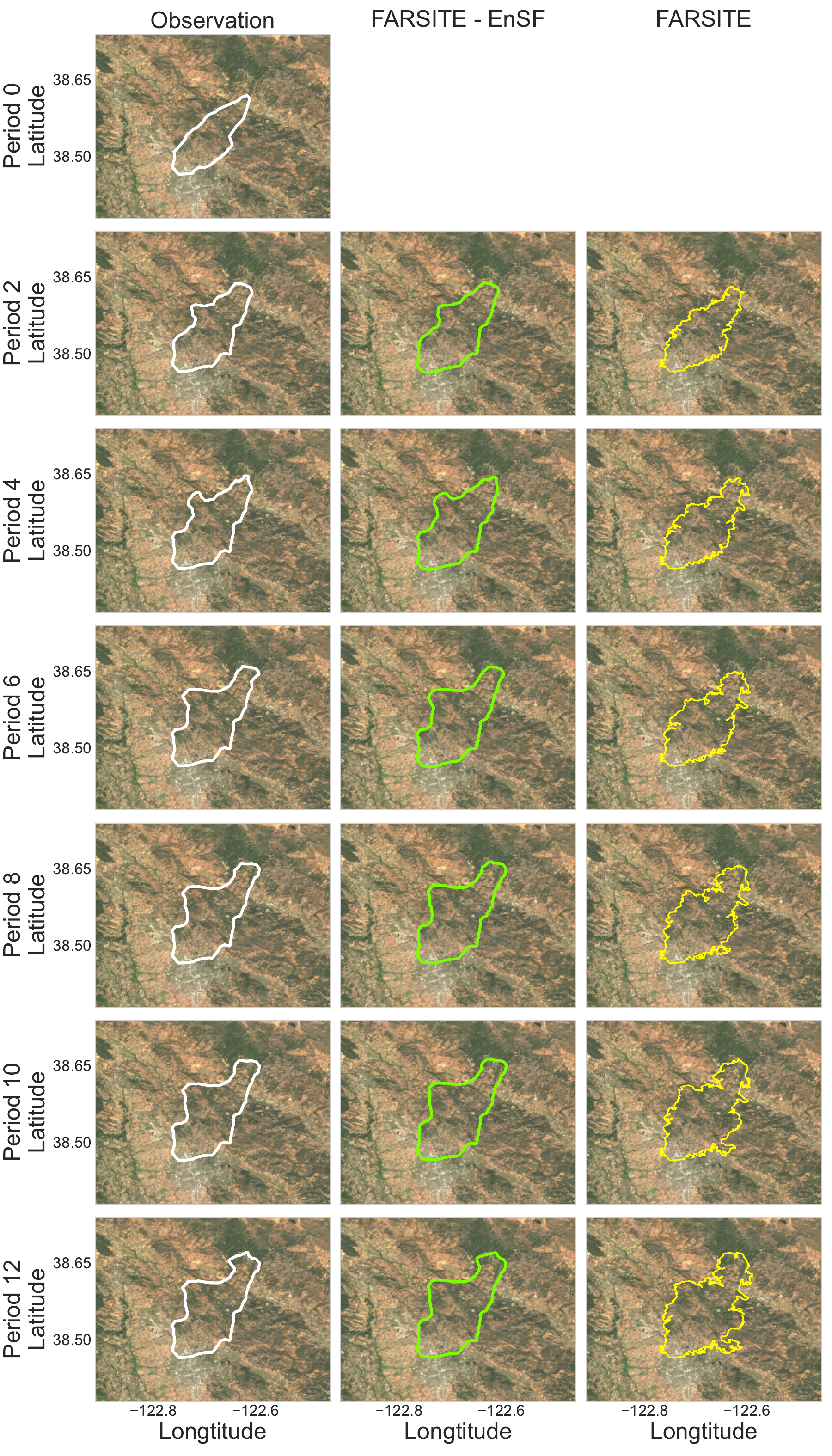}
    \caption{Comparison of fire perimeter estimations (0.5 days forecasting interval) using FARSITE and EnSF methods for fires 3173. Background imagery is sourced from LANDSAT.}
    \label{fig:3173}
\end{figure}

It is also noted that the simulation's forward function preserves point ordering after forecasting and filtering, but FARSITE operates differently by only expanding the current fire perimeter. Larger burned areas require more vertices to define their boundaries, creating critical challenges for EnKF. When EnKF cannot filter effectively at certain steps, it fails to generate polygons with consistent clockwise or counterclockwise ordering. Point matching was performed clockwise based on the original configuration, with the starting point closest to 0 radians. This ordering disruption confuses FARSITE when areas become multiply covered, significantly increasing EnKF computation time. In Figures \ref{fig:3128} and \ref{fig:3173}, we present the comparison results for the two larger fires 3128 and 3173, and the observations are consistent. 

\clearpage

\section{Conclusions and Discussion}
\label{sec:5}
This paper provided a comprehensive investigation of the data assimilation problems for fire perimeter prediction using the EnSF. Technical details have been provided, and the numerical experiments successfully demonstrated that the EnSF can be used as an alternative tool to traditional wildfire data assimilation methods for the FARSITE and other numerical fire prediction models. We also noted several limitations in this study that open avenues for future research. We only use single-polygon fire perimeters for both forecasts and observations, where this approach avoids a significant methodological challenge: developing a robust vertex-matching method to compare topologically different shapes, such as when a multi-polygon observation must be compared to a single-polygon forecast. Addressing this complex point-matching problem is a critical direction for future work. Furthermore, subsequent research should focus on integrating dynamic fuel moisture data to improve model realism and accuracy.

\bibliographystyle{abbrv}
\bibliography{paper}

\appendix

\section*{Appendix: Implementation, Code and Data}
\label{app:code}

In the Appendix, we provide details on implementation, code, and data. 

\textbf{Repository and Dataset Access.} The code and data are available on GitHub at \url{https://github.com/Song-Platelet/Fire-EnSF}. The fire boundary dataset can be accessed through \url{https://springernature.figshare.com/collections/A_Dataset_of_California_Wildfire_Spread_Derived_Using_VIIRS_Observations_and_an_Object-based_Tracking_System/5601537/1}. 
The users need to download the \texttt{Largefire.tar.gz} file and extract LargeFires\_2012-2020.gpkg, which contains the fire boundary information.

\textbf{FARSITE Integration.} This paper uses the C/C++ version of FARSITE. To integrate FARSITE with the EnKF and EnSF frameworks, Python calls the executable through the command prompt with the \texttt{subprocess} module. 

\textbf{Installation and Setup. }
To start the code, first download FARSITE from GitHub. For Linux systems, run \texttt{farsite\_import.py} to obtain FARSITE. For Windows systems, use the version in the \texttt{src} folder, which contains revisions to replace Linux-specific functions with Windows-compatible alternatives. The makefile has been updated accordingly. The prediction results show no significant difference between Linux and Windows versions.

For Windows compilation, install the MinGW toolkit following this tutorial: \url{https://www.youtube.com/watch?v=oC69vlWofJQ}. After loading all C/C++ files, run \texttt{mingw32-make all} to create \texttt{TestFARSITE.exe}.

Note that FARSITE primarily focuses on fire behavior modeling, so fire boundary forecasting is only one of its functions. The model generates multiple outputs beyond fire boundaries. Our tests show that FARSITE spends most execution time creating output files. We recommend removing all output file generation except for the forecasted fire boundary to save computational time. The Windows version already implements these modifications.

\textbf{Data Processing.} The file \texttt{data\_process.ipynb} contains code for data collection and cleaning. While the processed data is available on GitHub, many fires remain unexplored and offer opportunities for experimentation with different fire events.

This code was developed in March 2025. The \texttt{landfire} module can be downloaded and imported in Python; however, current data requests fail and return JSON errors. We suggest two alternatives: first, request data directly from LANDFIRE, which will deliver data via email; second, create custom landscape files with all layers as described in Table \ref{tab:farsite_landscape}.

Weather data was obtained from Google Earth Engine (GEE) \cite{noaagfs0p25gee}. GEE requires registration at \url{https://code.earthengine.google.com/register}. For detailed access instructions, refer to the official Google Developers guide at \url{https://developers.google.com/earth-engine/guides/access}.

Meanwhile, we encountered a problem with the \texttt{osgeo} package on Linux --- it wouldn't install properly. Because of this, we were only able to run \texttt{data\_process.ipynb} on Windows, where the package installed successfully.

\textbf{Code Structure.} The file \texttt{util.py} contains all commonly used functions, such as RMSE computation and FARSITE execution.
While some functions in \texttt{Data\_Process.ipynb} are currently non-functional, the paper's data is available on GitHub. There are four folders with four-digit numbers that correspond to fire IDs. Each folder contains the landscape file and its projection file. Other folders are named by period, where 0 indicates the starting point for FARSITE manipulation. Each period folder contains several files: files starting with \texttt{fire\_boundary} are shapefiles of the fire boundary at the current period with corresponding files, and \texttt{para.input} is the input file generated from \texttt{Data\_Process.ipynb}.

\textbf{Simulation Scripts.} The file \texttt{sim\_fine\_tune.ipynb} simulates the simple dynamic system described in Equation \ref{eq:sim_set} for both coarse tune and fine tune results, as shown in Figures \ref{fig: sim_tune} and \ref{fig:sim_ftune}. The script \texttt{sim\_run.py} implements the simple dynamic system with data assimilation, EnSF, and EnKF as described in Section \ref{subsec:sim}.

This implementation primarily focuses on the comparison between fine-tuned EnSF and EnKF. The script saves an RMSE comparison plot as \texttt{num\_com.pdf} as illustrated in Figure \ref{fig:sim_com}. To evaluate computational consumption, the script executes \texttt{trials} iterations for both EnSF and EnKF, and records the execution time in a DataFrame. The descriptive statistical report of time records is saved to \texttt{time\_com\_result.txt}.

\textbf{Application Scripts.} \texttt{farsite\_run.py} implements FARSITE without data assimilation methods. For each period $i$, the input fire boundary is the last estimated fire boundary (the initial input uses the observation, so RMSE starts at 0 for period 0). The instruction file is saved in the period folder as \texttt{run.txt}. After execution, FARSITE saves output files in the \texttt{out} folder within the current period folder. Since this FARSITE version forecasts fire boundaries hourly, the output file \texttt{output\_Perimeters.shp} contains hourly fire boundaries as linestrings for the 12-hour estimation. Our study requires only the final estimation, which we save as \texttt{output.shp} in the \texttt{out} folder for verification purposes.

\texttt{farsite\_ensf.py} and \texttt{farsite\_enkf.py} implement FARSITE with EnSF and EnKF respectively. These scripts share similar approaches for the prediction step. They create an \texttt{ensemble} folder under each period, which contains individually named folders for each ensemble member. Similar to \texttt{farsite\_run.py}, FARSITE predictions are saved in the \texttt{out} folder, with the final fire boundary saved as \texttt{output.shp}. Since each ensemble member requires FARSITE prediction, we run FARSITE in parallel for multiple input files. For each period $i$, the estimated state saves as \texttt{ensf\_$i$.npy} or \texttt{enkf\_$i$.npy}. The EnSF implementation additionally requires fine-tuning, so it generates an \texttt{ensf\_final\_rmse.db} SQLite database with columns for fire ID, two hyperparameters, and average RMSE to record the tuning data.

Moreover, we used \texttt{GeoPandas} to read \texttt{.shp} files during the experiment. \texttt{GeoPandas} uses \texttt{fiona} to read these files as default. Still, we encountered a compatibility issue on Linux: the latest \texttt{fiona} version has updated code that \texttt{GeoPandas} has not yet adapted to. To fix this, use \texttt{pyogrio} instead by specifying \texttt{engine='pyogrio'} in the \texttt{read\_file} function.

\textbf{Data Visualization.} The script \texttt{farsite\_ensf.py} generates plots for dimension change, fine tuning in application, and fire perimeter comparison as demonstrated in Figures \ref{fig:dim}, \ref{fig:app_tune}, \ref{fig:1819}, \ref{fig:2286}, \ref{fig:3128}, and \ref{fig:3173}. The script saves these plots as separate files. Note that the fine-tuning plots and fire perimeter comparisons are specific to each fire, so they save under their corresponding fire ID folders.
\end{document}